# A new selection strategy for selective cluster ensemble based on Diversity and Independency

Muhammad Yousefnezhad[a], Ali Reihanian[b], Daoqiang Zhang[a] and Behrouz Minaei-Bidgoli[c]

[a]Department of Computer Science, Nanjing University of Aeronautics and Astronautics, China.
[b] Department of Electrical and Computer Engineering, University of Tabriz, Iran.
[c] Department of Computer Engineering, Iran University of Science and Technology, Iran.

**Abstract**

This research introduces a new strategy in cluster ensemble selection by using Independency and Diversity metrics. In recent years, Diversity and Quality, which are two metrics in evaluation procedure, have been used for selecting basic clustering results in the cluster ensemble selection. Although quality can improve the final results in cluster ensemble, it cannot control the procedures of generating basic results, which causes a gap in prediction of the generated basic results' accuracy. Instead of quality, this paper introduces Independency as a supplementary method to be used in conjunction with Diversity. Therefore, this paper uses a heuristic metric, which is based on the procedure of converting code to graph in Software Testing, in order to calculate the Independency of two basic clustering algorithms. Moreover, a new modeling language, which we called as "Clustering Algorithms Independency Language" (CAIL), is introduced in order to generate graphs which depict Independency of algorithms. Also, Uniformity, which is a new similarity metric, has been introduced for evaluating the diversity of basic results. As a credential, our experimental results on varied different standard data sets show that the proposed framework improves the accuracy of final results dramatically in comparison with other cluster ensemble methods.

## 1. Introduction

Clustering, one of the main tasks in data mining is to discover meaningful patterns in the non-labeled data sets (Fred and Lourenço, 2008; Strehl and Ghosh, 2002; Topchy et al., 2003). Generally, basic clustering algorithms cannot recognize accurate patterns in a complex data set because they optimize final clustering results according to their objective functions. In other words, patterns of each data set are recognized by a special perspective, according to the algorithm's objective function instead of natural relations between data points in each data set (Jain et al., 2004). Combining the primary clustering results which are generated by basic clustering algorithms will cause cluster ensemble to achieve better final results. There are two steps in cluster ensemble: in the first step, different results are generated from the basic clustering methods by using different algorithms and changing the number of partitions. In the second step, basic results (ensemble committee) are combined by using an aggregating mechanism which leads to the generation of the final result (Alizadeh et al., 2011, 2014; Alizadeh et al., 2012; Strehl and Ghosh, 2002). The second step is performed by consensus functions.

Introduced by Fern and Lin, 2008, selective cluster ensemble is a new approach which combines a *selected group* of best primary results according to consensus metric(s) from ensemble committee in order to improve the accuracy of final results. The selection strategy aims to select better partitions of ensemble committee. In recent years, Diversity and Quality have been used to select the basic clustering results. A proper selection strategy can reflect the implicit features of data sets, and the clustering performance can be improved (Alizadeh et al., 2011, 2014; Alizadeh et al., 2012; Fern and Lin, 2008; Jia et al., 2012; Limin and Xiaoping, 2012). Quality cannot control the procedures of generating basic results and the prediction of the same results' accuracy. In order to evaluate the basic results according to the process of basic clustering algorithms, Yousefnezhad (2013) introduced Independency metric in D&I[1] method. With the same idea, Alizadeh et al., 2015 introduced a new method which is called WOCCE[2] in which the Independency is used as a criterion to map WOC[3], a theory in social science, to Cluster Ensemble Selection. Although, the performances and processes of D&I and WOCCE are the same (Alizadeh et al., 2015; Yousefnezhad et al., 2013), Alizadeh et al., 2015 used the concepts of WOC for heuristic proving of each components in D&I. In addition, they integrated the main assumptions of D&I into a new metric which is called "Decentralization". In the process of D&I and WOCCE, two algorithms which are of two different types are considered to be completely independent, and the Independency degree of two algorithms which are the same type is calculated by a random values matrix of those algorithms. For instance, the random values of k-means are the random values of clusters' centers in the first iteration of the algorithm (Alizadeh et al., 2015; Yousefnezhad et al., 2013). Since the performance and many concepts of D&I and WOCCE are the same, only D&I is used in this paper.

This paper proposes a new method for calculating Independency which is based on the procedure of converting code to graph in *"Software Testing"*. In this method both the same type algorithms and the algorithms which are in different types could have the Independency degree. The Independency degree is a value between zero and one which shows the probability of the generated result's accuracy based on analyzing the problem solving procedures of the algorithms. In addition, a new modeling language named as CAIL[4], is introduced in this paper which normalizes clustering algorithms' codes and pseudo codes. Moreover, this paper proposes a new metric based on APMM[5] for evaluating the diversity of basic results. Also, a new method for combing basic results which is based on EAC[6] (Fred and Jain, 2005), which is called WEAC[7], is introduced in this paper. The main contributions of this paper are:

1. In this paper, a new strategy for evaluating and selecting the best basic results in Cluster Ensemble Selection is introduced. This new strategy is based on the Independency and Diversity metrics.
2. Unlike the previous calculation of Independency which considered the two same type algorithms to be completely independent, this paper introduces a new method for calculating the real value of Independency degree between two same type algorithms. Also, this method can calculate the Independency degree between two different types of algorithms using the previous calculation of Independency.
3. For evaluating Independency metric, this paper introduces a new modeling language named as CAIL which is designed for estimating Independency degree in Clustering problems.

---

[1] Diversity and Independency
[2] Wisdom Of Crowds Cluster Ensemble
[3] The Wisdom of Crowds
[4] Clustering Algorithms Independency Language
[5] Alizadeh-Parvin-Moshki-Minaei
[6] Evidence Accumulation Clustering
[7] Weighted Evidence Accumulation Clustering

4. This paper introduces Uniformity which is a greedy metric for evaluating diversity of two basic results. This metric is based on APMM.
5. This paper introduces WEAC which is a new method for combining weighted basic results based on EAC. While this paper uses Independency degree as a weight in WEAC for generating final clustering result, any other metric can be used as a weight in WEAC for different clustering solutions in future works.

The Main goals of this paper are to improve the performance of D&I (Yousefnezhad, 2013; Yousefnezhad et al., 2013), or WOCCE (Alizadeh et al., 2015) by proposing the new calculation of Independency and Diversity metrics; and also to omit the thresholding procedures of Independency in the two mentioned methods (D&I & WOCCE).

This paper is organized as follows. Section 2 describes previous works on selective cluster ensemble. Section 3 presents our proposed method. In Section 4, our experimental results on 17 different scaled standard data sets are presented. Finally, conclusions are given in Section 5.

## 2. Background

### 2.1. Clustering analysis

The major aim of data clustering is to find groups of patterns (clusters) in such a way that patterns in one cluster can be more similar to each other than to patterns of other clusters (Akbari et al., 2015). A clustering algorithm decides each input data belongs to which cluster (Bahrololoum et al., 2015). Thus, Clustering can be considered as a powerful tool to reveal and visualize structure of data (Izakian et al., 2015). Basic clustering algorithms optimize the final clustering results according to their objective functions. In other words, patterns of each data set are recognized by a special perspective according to the objective functions of algorithms instead of natural relations between data points in each data set. Analyzing similarity and properties of clustering algorithms' objective functions is necessary for generating best results in cluster ensemble selection. Jain et al., 2004 proposed taxonomy of clustering algorithms according to their objective functions. They proved that the methods which are in a group (with the same objective function) have almost the same performances on a particular data set. Moreover, many algorithms can be found which are developed based on a specific algorithm such as, algorithms which are the extension of k-means (Jain, 2010), or linkages (Gose, 1997). These facts motivate researchers to propose cluster ensemble methods.

Cluster ensemble proved that better final results can be generated by combining basic results instead of only choosing the best one. Generally, a cluster ensemble has two important steps (Jain et al., 1999; Strehl and Ghosh, 2002):
1. Generating different results from primary clustering methods using different algorithms and changing the number of their partitions. This step is called generating diversity or variety.
2. Combining the primary results and generating the final ensemble. This step is performed by consensus functions (aggregating mechanism).

It is clear that an ensemble with a set of identical models doesn't have any advantages. Thus, the aim is to combine models which predict different outcomes. In order to achieve this goal, there are four components to be changed which are data set, clustering algorithms, evaluation metrics, and combine methods. A set of models can be created from two approaches: Choosing data representation, and Choosing clustering algorithms or algorithmic parameters.

Strehl and Ghosh, 2002 proposed the Mutual Information (MI) for measuring the consistency of data partitions; Fred and Jain, 2005 proposed Normalized Mutual Information (NMI), which is independent of cluster size. This metric can be used to evaluate clusters and the partitions in many applications. For instance, Zhong and Ghosh, 2005 used NMI for evaluating clusters in document clustering and Kandylas et al., 2008 used it for community knowledge analysis. Fern and Lin, 2008 developed a method which effectively uses a selection of basic partitions to participate in the ensemble, and consequently in the final decision. They also used the Sum NMI (SNMI) and Pairwise NMI as quality and diversity metrics between partitions, respectively. Jia et al., 2012 proposed SIM for diversity measurement which works based on the NMI. Azimi and Fern, 2009 used cluster ensemble selection to avoid consensus partitions which are excessively different from the base partitions they result from. They demonstrated that their method can result in partitions with enhanced SNMI. Limin and Xiaoping, 2012 used Compactness and Separation for choosing the reference partition in cluster ensemble selection. They also used new diversity and quality metrics as a selection strategy. Alizadeh et al., 2011, 2014 and Alizadeh et al., 2012 explored the disadvantages of NMI as a symmetric criterion. They used the APMM and MAX metrics to measure diversity and stability, respectively, and suggested a new method for building a co-association matrix from a subset of base cluster results. This paper introduces Uniformity for diversity measurement, which works based on the APMM metric.

Algorithm's Independency degree in cluster ensemble selection is introduced in (Alizadeh et al., 2015; Yousefnezhad, 2013; Yousefnezhad et al., 2013). In their method, the primary clustering algorithms with different types are considered to be completely independent. Furthermore, the Independency degrees of clustering algorithms with the same types are calculated by "BPI" function. Also, Yousefnezhad et al., 2013 achieved final result by thresholding on generated basic results. Algorithm 1 shows BPI function's pseudo code (Yousefnezhad, 2013; Yousefnezhad et al., 2013).

**Algorithm 1:** Basic Primary Independency function (Yousefnezhad, 2013; Yousefnezhad et al., 2013)

---
**Function** BPI (C1, C2, P1, P2) **Return** [Result]
    **If** C1 and C2 are equal **Then**
        Distance-Matrix is distance between P1 and P2
        **Do until** Distance-Matrix is not null
            Find minimum cell of Distance-Matrix
            Store cell in Temp-Array
            Remove Row and Column of founded cell
            Create new Distance-Matrix
        **End loop**
        **Return** Result = Average of Temp-Array
    **Else**
        **Return** Result = 1 for depicting two algorithms are independent
    **End If**
**End Function**

---

In Algorithm 1, C1 and C2 represent the types of clustering algorithms. According to Algorithm 1, BPI returns "Result = 1" when the algorithms are of two different types. Indeed, BPI considers each two different types of algorithms to be fully independent. Also in Algorithm 1, P1 and P2 are basic parameters of the algorithms such as the initial seed points in k-means. Actually, any random values or parameters which can change the final result in basic clustering algorithms can be represented by P1 and P2 (Yousefnezhad, 2013; Yousefnezhad et al., 2013).

In this paper, two parts of the approach which is introduced by Yousefnezhad et al., 2013, have been improved. First, in order to model and evaluate the Independency of clustering algorithms, a new technique which is based on algorithms' graph codes is presented. Second, the algorithms' Independency degrees are used as weights to evaluate diversity in the process of generating the final result. After modifying these two parts, the thresholding for Independency metric (Yousefnezhad, 2013; Yousefnezhad et al., 2013) in the process of cluster ensemble selection is omitted. As a result, there will be no initial value for "*iT*" (independency Threshold) parameter as an input of the algorithm which will be proposed in the next section. In addition, in this algorithm, the Independency degree between each two basic clustering algorithms will be calculated by graph-based modeling.

**2.2. Graph software testing**

Software testing is an important part of software development to which almost 60% of the total production cost is assigned. "Software modeling", one of the main tasks in software testing, can be implemented with the help of syntax, input space, logic, or graph. A graph-based modeling can provide a graphical representation of the source code, software design, use cases, and etc. (Ammann and Offutt, 2008). It can be a useful mechanism for evaluation of procedures in clustering algorithms. This paper introduces Clustering Algorithms Independency Language (CAIL) which is a new modeling language for normalizing codes and pseudo codes in which the concepts of graph-based modeling are used for calculating the degree of Independency for basic clustering algorithms. Also, a new instruction, which is based on the requirements in Independency evaluation, is proposed for transforming CAIL codes into graphs.

**3. Proposed method**

This section introduces a supplementary method, using diversity and Independency metrics, for selecting best partitions in an ensemble committee. Figure 1 illustrates our proposed framework.

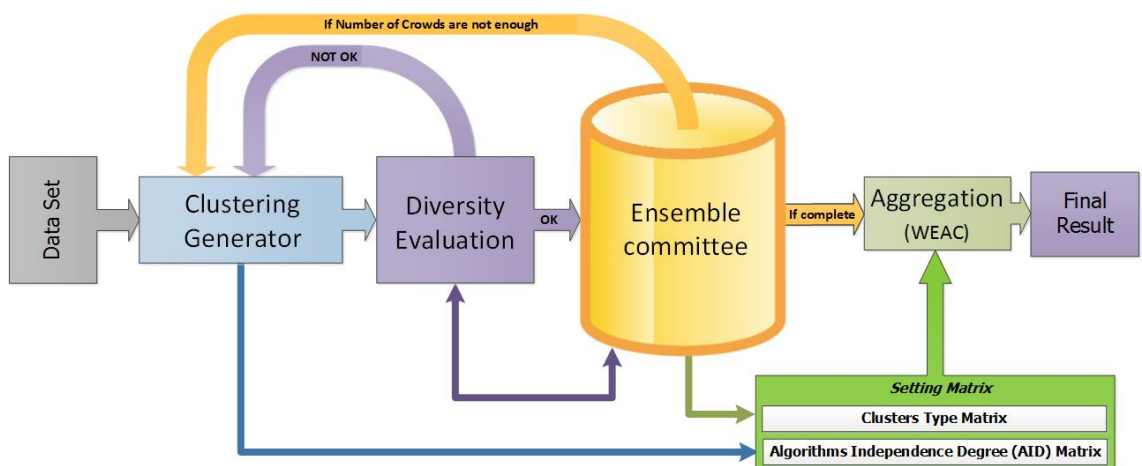

**Fig. 1.** The framework of the proposed method

Figure 1 shows how a final result is generated in our proposed method. Generally, it can be said that in the proposed method, a data set is divided into non-aligned clusters in 3 stages; in the first stage, a basic clustering algorithm generates a result from the data set. In the second stage, this generated result is evaluated by diversity metric and the evaluated result is added to ensemble committee only if it has an acceptable diversity degree. The above two stages are repeated until the number of ensemble committee members reach to enough amount. Then, the final result is created by using the members of ensemble committee and their independence degrees.

The rest of this section is organized as follows: First, diversity metric is introduced. After that, the concept of Independency in clustering algorithms is explained. Then, a new method for transforming the clustering algorithm's codes and pseudo codes into graphs is presented. Next, CAIL code analyzer, software for automatically comparing the Independency of clustering algorithms, is introduced. After that, the pseudo code of our proposed method is presented. Finally, the summary of the proposed method is given.

### 3.1. Diversity

After generating individual clustering results in Cluster Ensemble Selection methods, a consensus function must be used to evaluate them. NMI is used as the consensus function by most of the classical methods. Since NMI is a symmetric method, Alizadeh et al., 2011, 2014 and Alizadeh et al., 2012 concluded the disadvantages of it and therefore, they proposed APMM and MAX for solving the symmetry problem in the NMI. The APMM is calculated as follows (Alizadeh et al., 2011, 2014):

$$APMM(C,P) = \frac{-2n_c \log\left(\frac{n}{n_c}\right)}{n_c \log\left(\frac{n_c}{n}\right) + \sum_{i=1}^{kp} n_i^p \log\left(\frac{n_i^p}{n}\right)} \quad (1)$$

In Eq. 1, $n_c$, $n_i^p$, and n are the size of cluster C, the size of the i-th cluster of partition P, and the number of samples which are available in the partition of cluster C, respectively. $k_p$ is the number of clusters in the partition P. As a matter of fact, the only difference between NMI and APMM is that the first one (NMI) compares two partitions while the second one (APMM) compares a partition with a cluster.

To calculate the similarity of partition P with respect to a partition of the reference set (ensemble committee), this paper uses AAPMM which is calculated as follows (Alizadeh et al., 2011, 2014):

$$AAPMM(P,P^*) = \frac{1}{N}\sum_{i=1}^{N} APMM(C_i, P^*) \quad (2)$$

In Eq. 2, $P^*$ is a partition from reference set, $C_i$ is the *i-th* cluster of partition P, and N is the number of clusters in the partition P.

This paper proposes a redefined version of APMM, because the original version only measures the diversity between a cluster in the first partition and all of clusters in the second partition (Alizadeh et al., 2014). This redefined metric which is called Uniformity is used for evaluating the diversity between a partition and a reference set as ensemble committee. In other words, this metric is used to satisfy the Diversity criterion in the proposed method. The Uniformity is defined as follows:

$$\text{Uniformity}(P) = \max_{i=1}^{n}(AAPMM(P,P_i)) \quad (3)$$

In Eq. 3, $P_i$ is the i-th partition in ensemble committee. n is the number of members in reference set. Uniformity represents the maximum value of similarity between partition P and the other partitions of ensemble committee. Since Uniformity is normalized between zero and one, we consider 1 – Uniformity to represent the diversity as follows:

$$DIV(P) = 1 - \text{Uniformity}(P) \tag{4}$$

As mentioned before, one of the conditions that should be established in order to append a partition to the ensemble committee (which is known as the diversity condition) is as follows:

$$DIV(P) \geq dT \tag{5}$$

This means that if the diversity of a generated partition satisfies *dT* (diversity threshold), it will be added to the reference set.

### 3.2. Independency

Before this paper starts to explain the details of Independency, two questions must be answered: First, it should be clear what the main goal of using Independency is. In the proposed method, the correctness of the generated individual clustering results can be estimated by Independency. As a matter of fact, Independency tries to estimate the correctness by comparing the similarity of clustering algorithms in the process of solving a clustering problem. In other words, this paper considers the correctness of two same (low-value in the diversity estimation) individual clustering results. The individual clustering results are considered to be low, when they are generated by the clustering algorithms with similar objective functions; on the other hand, the correctness of two same individual clustering results are considered to be high, when they are generated by two clustering algorithms with different objective function even if those results don't have a significant diversity. This comparison is considered to be reliable for complex data sets, practically. Indeed, in real-world data sets, there is no class-label. Therefore, this is one of the best ways for estimating the correctness of the generated results, especially the same 'individual clustering results' (Fred and Lourenço, 2008). The second question is that how this technique can be used in software such as SAS or SPSS, in which the code of clustering algorithms cannot be find? For each clustering algorithm, the process of solving problem is unique. Therefore, if the implementations of an algorithm in two different programming languages are converted and normalized based on the proposed method of this paper, the results must be the same. Therefore, each open source codes of that algorithm can be used.

Figure 2 shows how an algorithm's graph array is generated for evaluating clustering algorithm's Independency degree. According to Figure 2, it can be said that a clustering algorithm's code is converted to graph array in 4 stages. First, Standard Code Mapping Table (SCMT), which is a consensus table, is prepared by looking in algorithms' codes. In other words, this table contains mathematical, statistical, heuristic, and other kinds of functions which are used in the algorithms. Also, this table, which is unique for each clustering problem, contains all mentioned functions which are used in the basic clustering algorithms. After that, the clustering algorithms' codes are manually converted to CAIL scripts with considering the SCMT table. Then, the algorithms' graphs are generated by using the CAIL codes which are generated in the previous stage. Finally, the weighted edges are stored in an array for evaluating algorithm's Independency degree. This array is called the graph's array.

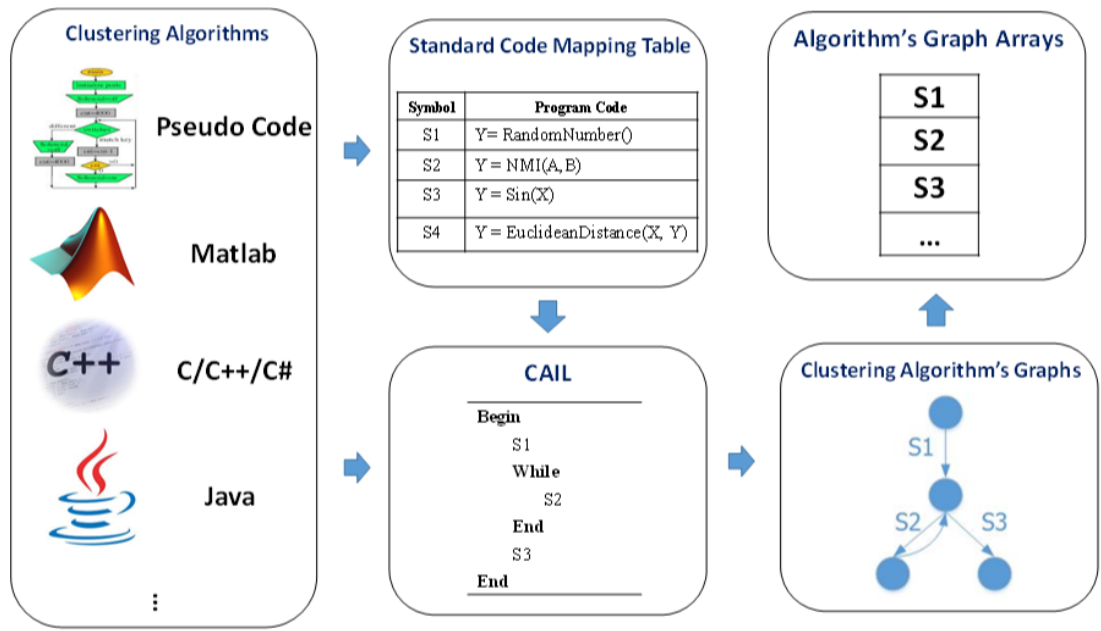

**Fig. 2.** The framework of the clustering algorithms' Independency evaluation

### 3.2.1. Clustering Algorithms Independency Language

In CAIL modeling, symbols are used instead of original codes or pseudo-codes of clustering algorithms. The main reasons are that: first, Codes or pseudo-codes are usually written in a standard language structure, so they need to be converted in a homogenous form in order to be compared with each other. What's more, the codes have many useless details such as different variables' definitions. Also, many mathematical equations and pseudo codes, used in algorithms, are not clear in papers.

This paper proposes a modeling method with considering SCMT's symbols. This method is not sensitive to implementation details. The procedure of converting codes to CAIL format is performed in five stages. These stages are listed as follows:

1. First, all additional codes such as the definitions of different variables and constants, descriptions, input and output commands, and each code that is not involved in the clustering process are omitted. Also, the implementations of the specific functions which are used in main function are omitted. For example, the implementation of evaluation metrics such as NMI, APMM and etc. can be omitted because they are shown in the SCMT table as symbols.
2. The logical operators in conditions and loops are removed because they do not affect the shape of the algorithm's graph.
3. All conditions, such as if, case, and etc., are converted to a unique format such as "if, else, end". Also, the loops like for, while, repeat, and etc., are converted to a unique format such as "while break end". Indeed, all formats of conditions and loops are used for quickly implementation of algorithms' codes by programmers. They are not important because algorithms' processes are implemented in clustering algorithm's Independency modeling instead of implementations of individual codes. These processes must affect the Independency. This paper uses "if, else, end" for all forms of conditions and "while, break, end" for all forms of loops.
4. The keyword "Begin" is added at the beginning and the keyword "End" is added at the end of a CAIL code for clarity of our definitions.

5. Generating SCMT table. This consensus table contains all mathematical, statistical, heuristic and other functions which are used in basic clustering algorithms. In order to name symbols in SCMT table, this paper recommends the following instructions: first, all functions should be grouped according to their types. Each group can be shown by a single English word. For instance, the R shows random function group, M shows mathematical function group and H shows heuristic function group. Each function can be shown by the name of its group along with a number in a bracket (see Table 3 as an example of the SCMT table).

Algorithm 2 and Algorithm 3 show two examples of CAIL scripts for k-means and FCM algorithms, respectively. These algorithms are generated according to the SCMT table which is illustrated in Table 3. According to these two algorithms, one of the advantages of the CAIL code is that it does not contain any implementation details. Also, another advantage of using CAIL and SCMT is that the codes and pseudo codes can be used together for modeling clustering algorithms.

| **Algorithm 2:** K-means in the CAIL format | **Algorithm 3:** FCM in the CAIL format |
|---|---|
| **Begin** | **Begin** |
|   R(1) |   R(1) |
|   **While** |   **While** |
|     F(1) |     M(2) |
|     M(1) |     M(3) |
|   **End** |   **End** |
| **End** | **End** |

### 3.2.2. Converting CAIL to Independency Graph

Independency graph is a special-purpose application graph. In addition, the CAIL codes model algorithms' Independency to a standard format. Thus, this paper does not use the same graph-based model structure, which is used in software testing, for converting CAIL to an algorithm's graph. However, this paper uses a custom format of graph-based modeling according to the evaluation of algorithms' Independency requirements. In this method, the code conjunctions, which are the "Begin", "End", conditions, loops and their sub-sectors are converted to nodes. The codes between each two nodes are considered as their edge. Like software testing approaches, this method uses directed graph for modeling algorithms. In the generated graphs in software testing, codes of each segment are written inside of its node. Also, the logical operation of conditions or loops is usually written on the edges. Unlike software testing, in our proposed method, the codes of each segment are placed on the corresponding edges. The main reasons for this can be mentioned as follows; first, the logical operations are omitted. Then, the evaluation process is important in this method. After that, the CAIL codes are pruned in the previous section. These pruned codes use standard codes according to the SCMT table. Finally, the process of each algorithm can clearly be visible in this status.

In our proposed method, the codes on each edge are considered as a non-numerical weight. An array of weighted edges, which is called the Independency graph's array, is used for storing the Independency graph in memory. Arrays are compared in order to calculate the Independency degree of the algorithms. Figure 3 and Figure 4 show two examples of CAIL codes and their converted graph.

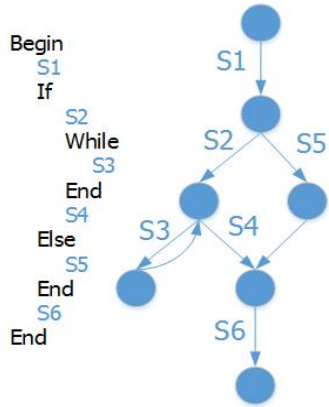 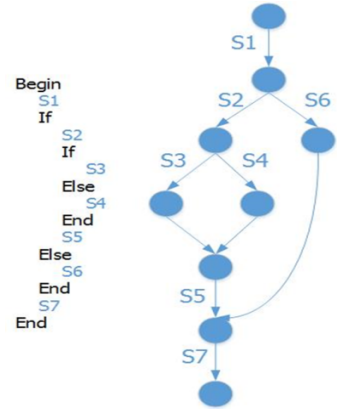

**Fig. 3.** An example of a CAIL code when it contains a loop and a condition

**Fig. 4.** An example of a CAIL code when it contains two conditions

### 3.2.3. Evaluating Independency Graph

Figure 5 shows the general structure of the Code Dependence Degree Matrix (CDDM) which is used for evaluating the Independency of two clustering algorithms.

$$CDDM = \begin{array}{c} \\ A\lg 2[1] \\ A\lg[2] \\ \vdots \\ A\lg 2[m] \end{array} \begin{array}{cccc} A\lg 1[1] & A\lg 1[2] & \cdots & A\lg 1[n] \\ \begin{bmatrix} a_{11} & a_{12} & \cdots & a_{1n} \\ a_{21} & a_{22} & \ddots & a_{2n} \\ \vdots & \vdots & \ddots & \vdots \\ a_{m1} & a_{m2} & & a_{mn} \end{bmatrix} \end{array}$$

**Fig. 5.** The Code Dependence Degree Matrix (CDDM)

According to Figure 5, each cell of CDDM matrix is calculated by the "Compare" function. Algorithm 4 gives the pseudo code of the "Compare" function. As this figure shows, this function compares the cells of Independency graph arrays. This figure shows how each cell of the first algorithm's array compared with all cells of the second algorithm's array. In this function, the "Count" variable is incremented if it can find one same symbol in the second algorithm's array for each symbol in the first algorithm's array. MSymbol represents the maximum number of symbols (blocks) in cell1 and cell 2. For instance, if cell 1 contains 5 blocks and cell 2 contains 6 blocks, the value of Maxsym will be 6. Final result is normalized by dividing the "Count" by MSymbol. The normalized value, which is called CDD, is stored in CDDM matrix cell which represents the intersection of two mentioned cells. This function finds the maximum value of CDDM's cells (the maximum values of CDDs), which is called the $MaxCell_i$, and stores them for calculating the Independency degree of an algorithm according to its corresponding CDDM matrix. After that, this function removes the MaxCell's row and column in CDDM matrix. In other words, for each block - each symbol in a cell is called a block - of the first algorithm, the most similar block in the second algorithm is found. From the second algorithm's blocks, the function finds the most similar block to the next block of the first algorithm by removing the row and columns of this block (MaxCell) from the CCDM matrix. Then, the function calculates the new MaxCell for new generated CDDM matrix. Finally, when the size of the CDDM matrix reaches to zero, this process is finished.

**Algorithm 4:** Compare Function

---
**Function** Compare (Cell1, Cell2) **Return** [CDD]
    Count = 0
    **While** we have Symbol in Cell1
        Sym1 = **Select** an Symbol in Cell1
        **Foreach** Sym2 in Cell2
            **If** Sym2 = Sym1 is found **Then**
                Count++
                Break
            **End If**
        **End Foreach**
    **End while**
    MSymbol = Max-Sym (Cell1, Cell2)
    **Return** CDD = Count / MSymbol
**End Function**

---

Eq. 6 shows the Algorithms' Independency Degree (AID) which is calculated at each step. In this equation, n is the minimum number of cells in the first and second algorithm's arrays, and m is the maximum number of cells in the first and second algorithm's arrays.

$$AID(A\lg_i, A\lg_j) = 1 - \frac{1}{m}\sum_{i=1}^{n} MaxCell_i \qquad (6)$$

The calculated results of Eq. 6 are stored in the Algorithm Independency Degree Matrix (AIDM) in order to be used in cluster ensemble selection. The size of AIDM matrix is $n \times n$ in which n is the number of algorithms in the cluster ensemble. The Eq. 7 shows how the AIDM cells are calculated.

$$a_{ij} = \begin{cases} AID(A\lg_i, A\lg_j) & i \neq j \\ -1 & i = j \end{cases} \qquad (7)$$

Since this paper uses BPI function (Yousefnezhad, 2013; Yousefnezhad et al., 2013) for calculating the Independency degree of algorithms with the same type, Eq. 7 assigns "-1" to Independency degree of each algorithm in comparison with itself. The final Independency degree is calculated by Eq. 8 during the running processes of algorithms.

$$AI[A\lg_i, A\lg_j] = \begin{cases} AIDM[A\lg_i, A\lg_i] & i \neq j \\ \frac{1}{m}\sum_{k=1}^{m} BPI[i,k] & i = j \end{cases} \qquad (8)$$

In Eq.8, Algi is a member of the selected algorithms in ensemble committee. m is the number of algorithms in the ensemble committee which are in the same type of Algi or Algj. Furthermore, AIDM is calculated by Eq. 7; and BPI is calculated by the pseudo code which is represented in Algorithm 1 (The basic parameters Independency function). Figure 6 illustrates an example of CDDM matrix for comparing k-means (K) and FCM (F) algorithms, which are defined by CAIL script in Algorithm 2 and Algorithm 3. Furthermore, the MaxCell values are {MaxCell1=1, MaxCell2=0}, and also AI[K,F] = AID[K,F] = 0.5 (see Figure 9).

$$CDDM = \begin{array}{c} K[R(1)] \\ K[F(1),M(1)] \end{array} \begin{bmatrix} F[R(1)] & F[M(2),M(3)] \\ 1 & 0 \\ 0 & 0 \end{bmatrix}$$

**Fig. 6.** The CDDM for comparing k-means and FCM based on Algorithm 2 and Algorithm 3

### 3.3. Weighted Evidence Accumulation Clustering

In order to select the evaluated individual results in cluster ensemble selection, thresholding is used. Then, with using the consensus function on the selected results, the co-association matrix is generated. At last, by applying linkage methods on the co-association matrix, the final result is generated. These methods generate the Dendrogram. After that, they cut the Dendrogram based on the number of clusters in the result (Alizadeh et al., 2015; Fred and Jain, 2005). In recent years, Evidence Accumulation Clustering (EAC) has been used in many researches as a high performance consensus function for combining individual results (Alizadeh et al., 2011, 2014; Alizadeh et al., 2012; Alizadeh et al., 2015; Azimi and Fern, 2009; Fern and Lin, 2008; Fred and Jain, 2005). EAC divides the number shared by objects over the number of partitions in which each selected pair of objects is simultaneously presented. EAC uses Eq. 9 for generating the co-association matrix.

$$C(i,j) = \frac{n_{i,j}}{m_{i,j}} \qquad (9)$$

In the above equation, $m_{i,j}$ is the number of partitions in which this pair of objects (i and j) is simultaneously presented and $n_{i,j}$ represents the number of clusters shared by objects with indices i and j. As a matter of fact, EAC considers that the weights of all algorithms' results are the same. This paper proposes Eq. 10 for generating the co-association matrix with considering the Independency degree of algorithms as a weight of combining the basic results. In this equation, AI is calculated by Eq. 8:

$$C(i,j) = \frac{\sum_{n_{i,j}} AI[A\lg_i, A\lg_j]}{m_{i,j}} \qquad (10)$$

Figure 7 shows the process of generating final result by using WEAC.

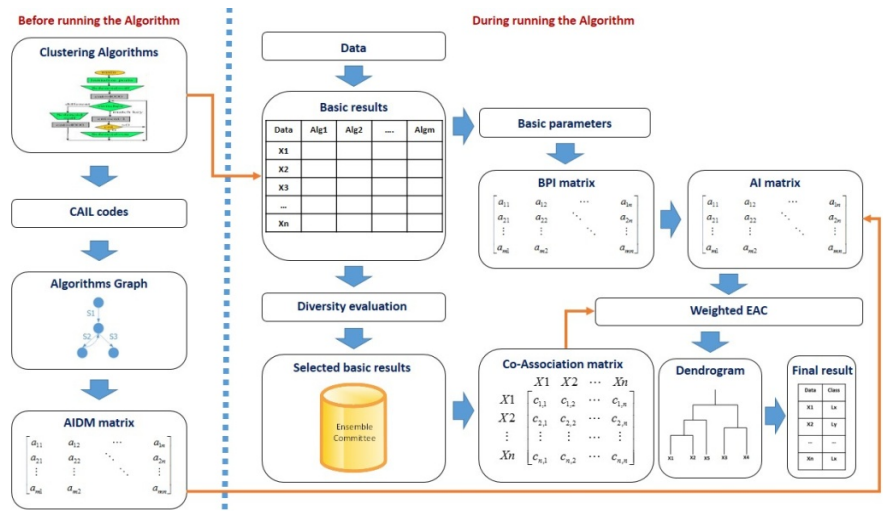

**Fig. 7.** The process of generating final result

As Figure 7 depicts, the process of generating the AIDM matrix is done before running the algorithm. Indeed, it decreases the runtime of the algorithm, which will be discussed later in section 4.2.

**3.4. Summary of the Proposed Method**

Algorithm 5 depicts the pseudo code of the proposed method. In Algorithm 5, Kb is the number of clusters in the final result, and dT is the diversity threshold. The distances are also measured by a Euclidean metric. The Generate-Basic-Algorithm function builds the partitions of base clusters (basic results), Generate-AI-Matrix builds the co-association matrix according to Eq. 9 by using the AIDM matrix and the results of BPI function. The Average-Linkage and Cluster functions build the final ensemble according to the Average Linkage method. The parameter Result is the final ensemble result, and nCE is the number of members in the ensemble committee.

**Algorithm 5:** The Proposed Method

---
**Function** CES (Dataset, Kb, dT) Return [Result, nCE]
    Initialize nCE to zero
    **While** we have base cluster
      [IDX, Basic-Parameter] = Generate-Basic-Algorithm (Dataset, Kb)
      **If** (Diversity (IDX) >dT) **then**
          Find the Algorithms AID from AIDM
          Insert idx, AID, and Basic-Parameter to Ensemble-Committee
          nCE = nCE + 1
      **End if**
    **End while**
    AI = Generate-AI-Matrix (AIDM, BPI)
    W-Co-Acc = WEAC (Ensemble-Committee, AI)
    Z = Average-Linkage (W-Co-Acc)
    Result = Cluster (Z, Kb)
**End Function**

---

There are three questions, which must be answered before this paper starts to explain the empirical results. First, "what is the main goal of using Independency estimation?". In the proposed method, Independency tries to estimate the correctness of generated individual clustering results by comparing the similarity of clustering algorithms in the process of solving a clustering problem. In other words, this paper considers the correctness of two same (low-value in the diversity estimation) individual clustering results to be low when they are generated by the clustering algorithms with similar objective function; and also, it considers the correctness of two same individual clustering results to be high when they are generated by two clustering algorithms with different objective functions even if those results don't have a significant diversity. In practice, this comparison can be reliable for complex data sets. Indeed, there is no class-label in real-world data sets; and this is one of the best ways for estimating the correctness of the generated results, especially the same "individual clustering results" (Alizadeh et al., 2015). The Next question to be answered is "how can we use this technique in applications such as SAS or SPSS, which do not have the code of clustering algorithms?". The process of solving problems for each clustering algorithm is unique. So, if one algorithm, which is implemented by two different programming language or even two different structures of implementation, is converted and normalized based on the proposed method, the results must be the same. Furthermore, the proposed "Compare" function can calculate the same results for two different structures of implementation because it only uses the contents of cells. For instance, the results are the same when you change the blocks ("THEN" and "ELSE") in the "IF" condition. As a result, we can use other open source codes for those algorithms from the Internet. The last question to be answered is that "which level of abstraction must be used for converting the codes to CAIL scripts?". As we can see in the examples (Algorithm 2, Algorithm 3 and Figure 8), the CAIL codes are generated based on general structures of clustering algorithms. We mostly desire to compare the objective functions, distance metrics, general processes of solving problem, and everything which can mathematically or technically change the performance of clustering results. Indeed, it is important to use a unique structure (and a common SCMT table) for all clustering algorithms in an individual clustering problem, while a given algorithm can be implemented in different ways. In fact, we can report the employed CAIL scripts like other parameters in the experiment, such as distance metric, types of basic clustering algorithms, etc.

## 4. Experiments

This section describes a series of empirical studies and reports their results. In real world, unsupervised methods are used to find meaningful patterns in non-labeled data sets such as web documents. Since real data sets don't have class labels, there is no direct evaluation method for evaluating the performance in unsupervised methods. Like many previous researches (Alizadeh et al., 2014; Alizadeh et al., 2012; Alizadeh et al., 2015; Fern and Lin, 2008; Fred and Jain, 2005; Yousefnezhad, 2013; Yousefnezhad et al., 2013), this paper compares the performance of its proposed method with other basic and ensemble methods by using standard data sets and their real classes. Although this evaluation cannot guarantee that the proposed method leads to high performances in all data sets in comparison with other methods, it can be considered as an example to demonstrate the superiority of the proposed method.

**Table 1** List of data sets and their related information

| No. | Name | Feature | Class | Sample |
|---|---|---|---|---|
| 1 | Half Ring | 2 | 2 | 400 |
| 2 | Iris | 4 | 3 | 150 |
| 3 | Balance Scale | 4 | 3 | 625 |
| 4 | Breast Cancer | 9 | 2 | 683 |
| 5 | Bupa | 6 | 2 | 345 |
| 6 | Galaxy | 4 | 7 | 323 |
| 7 | Glass | 9 | 6 | 214 |
| 8 | Ionosphere | 34 | 2 | 351 |
| 9 | SA Heart | 9 | 2 | 462 |
| 10 | Wine | 13 | 2 | 178 |
| 11 | Yeast | 8 | 10 | 1484 |
| 12 | Pendigits | 16 | 10 | 10992 |
| 13 | Statlog | 36 | 7 | 6435 |
| 14 | Optdigits | 64 | 10 | 5620 |
| 15 | Arcene | 10000 | 2 | 900 |
| 16 | CNAE-9 | 857 | 9 | 1080 |
| 17 | Sonar | 60 | 2 | 208 |

### 4.1. Data sets

The proposed method is applied to 17 different standard UCI data sets. Like many other papers and researches such as (Alizadeh et al., 2011, 2014; Alizadeh et al., 2012; Alizadeh et al., 2015; Yousefnezhad, 2013; Yousefnezhad et al., 2013), we have used the standard data sets to evaluate our numerous experiments. These standard data sets have no negative or positive effects on the performance of an algorithm. As a matter of fact, the reason of using the standard data sets is to conduct an evaluation with no artificial negative/positive bias and to compare different algorithms fairly.

We have chosen data sets which are as diverse as possible in their numbers of true classes, features, and samples, because this variety better validates the obtained results. These data sets are explained in Table 1. More information about these data sets is available in (Alizadeh et al., 2015; Azimi and Fern, 2009; Jain et al., 2004; Newman et al., 1998; Yousefnezhad, 2013; Yousefnezhad et al., 2013). The features of the data sets are normalized to a mean of 0 and variance of 1, i.e. N (0, 1).

### 4.2. CAIL code analyzer

As mentioned earlier, this paper develops an application for evaluating Independency degree by using CAIL codes. Figure 8 shows a snapshot of this application. This tool is developed by Microsoft C# .Net 2013. First, this application converts CAIL codes to graphs. After that, the graphs' arrays are stored in the memory. Finally, the arrays are compared with each other, and the Independency degree is shown in a message box. This application can work with any SCMT table that is prepared in formats which are described in section 3.2.1. As it is clear in Figure 8, two CAIL codes are given as inputs. In this figure, Code 1 implements K-means while Code 2 implements Spectral clustering using a sparse similarity matrix. Also, the message box represents the Independency degree of the two mentioned algorithms.

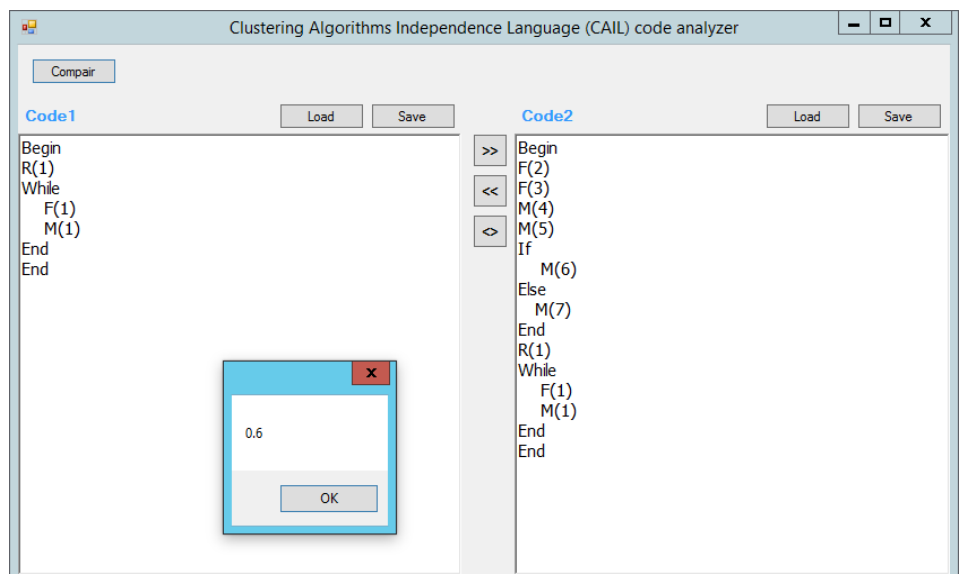

**Fig. 8.** The CAIL code analyzer

### 4.3. Performance Analysis

This paper used MATLAB R2014b (8.4) in order to generate experimental results. The algorithms which are described in Table 2 were used to generate the ensemble committee.

**Table 2** The standard code mapping table

| No. | Algorithm Name | ID |
|---|---|---|
| 1 | K-Means | K |
| 2 | Fuzzy C-Means | F |
| 3 | Median K-Flats | M |
| 4 | Gaussian Mixture | G |
| 5 | Subtract Clustering | SUB |
| 6 | Single-Linkage Euclidean | SLE |
| 7 | Single-Linkage Hamming | SLH |
| 8 | Single-Linkage Cosine | SLC |
| 9 | Average-Linkage Euclidean | ALE |
| 10 | Average-Linkage Hamming | ALH |
| 11 | Average-Linkage Cosine | ALC |
| 12 | Complete-Linkage Euclidean | CLE |
| 13 | Complete-Linkage Hamming | CLH |
| 14 | Complete-Linkage Cosine | CLC |
| 15 | Ward-Linkage Euclidean | WLE |
| 16 | Ward-Linkage Hamming | WLH |
| 17 | Ward-Linkage Cosine | WLC |
| 18 | Spectral clustering using a sparse similarity matrix | SPS |
| 19 | Spectral clustering using Nystrom method with orthogonalization | SPN |
| 20 | Spectral clustering using Nystrom method without orthogonalization | SPW |

Table 3 illustrates the SCMT table which is used in this paper.

**Table 3** The standard code mapping table

| No. | Symbol | Description |
|---|---|---|
| 1 | R(1) | Generate x random number |
| 2 | R(2) | Random Selection |
| 3 | M(1) | Y = EuclidianDistance(A, B) |
| 4 | M(2) | $c_j = \dfrac{\sum_{i=1}^{N} u_{ij}^{m} \times x_i}{\sum_{i=1}^{N} u_{ij}^{m}}$ |
| 5 | M(3) | $u_{ij} = \dfrac{1}{\sum_{k=1}^{C} \left( \dfrac{\|x_i - c_j\|}{\|x_i - c_k\|} \right)^{\frac{2}{m-1}}}$ |
| 6 | M(4) | Do exp function: $S = e^{(-1 \times (A^2 / 2 \times \sigma^2))}$ |
| 7 | M(5) | Do laplacian function: $L = D^{-\frac{1}{2}} \times S \times D^{-\frac{1}{2}}$ |
| 8 | M(6) | Largest magnitude |
| 9 | M(7) | Smallest magnitude |
| 10 | M(8) | Normalizing A and B |
| 11 | M(9) | Y=HammingDistance(A, B) |
| 12 | M(10) | Y=CosinDistance(A, B) |
| 13 | M(11) | $D_{SL}(C_i, C_j) = \min\limits_{a \in c_i, b \in c_j} d(a, b)$ |

| 14 | M(12) | $D_{cl}(C_i, C_j) = \max\limits_{a \in C_i, b \in C_j} d(a,b)$ |
|---|---|---|
| 15 | M(13) | $D_{AL}(C_i, C_j) = \dfrac{1}{n_i n_j} \sum\limits_{a \in n_i, b \in n_j} d(a,b)$ |
| 16 | M(14) | $D_{WL}(C_i, C_j) = \sqrt{\dfrac{2 n_i n_j}{(n_i, n_j)} \|\bar{a} - \bar{b}\|_2}_{a \in n_i, b \in n_j}$ |
| 17 | M(15) | $Q(\theta', \hat{\theta}^{(j)}) = E(\ell_0(\theta'; t) \mid z, \hat{\theta}^{(j)})$ |
| 18 | M(16) | $(\ell^*, m^*) = \arg\min\limits_{(\ell, m)} d(\ell, m)$ |
| 19 | M(17) | Z=X/Y |
| 20 | M(18) | $d_x P_i = -\dfrac{(P_i x x^T - P_i x x^T P_i^T P_i)}{\sqrt{1 - \|P_i x\|^2}}$ |
| 21 | M(19) | $i^* = \arg\max\limits_{1 \leq i \leq K} \|P_i x\|$ |
| 22 | F(1) | Assign each object to closest centroid/subspace |
| 23 | F(2) | Generate (t-nearest-neighbor) sparse distance matrix |
| 24 | F(3) | Convert distance matrix to similarity matrix |
| 25 | F(4) | Do orthogalization |
| 26 | F(5) | Restore cluster labels in orginal order |
| 27 | F(6) | Compute the proximity matrix |
| 28 | F(7) | Merge two closest cluster |
| 29 | F(8) | Y=Subclass(X) |
| 30 | F(9) | Update $P_{i^*} : P_{i^*} \mapsto P_{i^*} - dt d_x P_{i^*}$ |

Figure 9 shows AIDM matrix calculated by the SCMT table, which is described in Table 3, and the CAIL code analyzer.

|   | K | F | M | G | SUB | SLE | SLH | SLC | ALE | ALH | ALC | CLE | CLH | CLC | WLE | WLH | WLC | SPS | SPN | SPW |
|---|---|---|---|---|---|---|---|---|---|---|---|---|---|---|---|---|---|---|---|---|
| K | -1 | 0.5 | 0.65 | 1 | 0.4 | 0.75 | 1 | 1 | 0.75 | 1 | 1 | 0.75 | 1 | 1 | 0.75 | 1 | 1 | 0.6 | 0.67 | 0.67 |
| F | 0.5 | -1 | 0.9 | 1 | 0.9 | 1 | 1 | 1 | 1 | 1 | 1 | 1 | 1 | 1 | 1 | 1 | 1 | 0.8 | 0.84 | 0.84 |
| M | 0.65 | 0.9 | -1 | 1 | 0.55 | 1 | 1 | 1 | 1 | 1 | 1 | 1 | 1 | 1 | 1 | 1 | 1 | 0.86 | 0.88 | 0.88 |
| G | 1 | 1 | 1 | -1 | 0.94 | 0.84 | 0.84 | 0.84 | 1 | 1 | 1 | 0.84 | 0.84 | 0.84 | 1 | 1 | 1 | 1 | 1 | 1 |
| SUB | 0.4 | 0.9 | 0.55 | 0.94 | -1 | 0.65 | 0.9 | 0.9 | 0.75 | 1 | 1 | 0.65 | 0.9 | 0.9 | 0.75 | 1 | 1 | 0.76 | 0.8 | 0.8 |
| SLE | 0.75 | 1 | 1 | 0.84 | 0.65 | -1 | 0.25 | 0.25 | 0.25 | 0.5 | 0.5 | 0.25 | 0.5 | 0.5 | 0.25 | 0.5 | 0.5 | 0.9 | 0.92 | 0.92 |
| SLH | 1 | 1 | 1 | 0.84 | 0.9 | 0.25 | -1 | 0.25 | 0.5 | 0.25 | 0.5 | 0.5 | 0.25 | 0.5 | 0.5 | 0.25 | 0.5 | 1 | 1 | 1 |
| SLC | 1 | 1 | 1 | 0.84 | 0.9 | 0.25 | 0.25 | -1 | 0.5 | 0.5 | 0.25 | 0.5 | 0.5 | 0.25 | 0.5 | 0.5 | 0.25 | 1 | 1 | 1 |
| ALE | 0.75 | 1 | 1 | 1 | 0.75 | 0.25 | 0.5 | 0.5 | -1 | 0.5 | 0.5 | 0.25 | 0.5 | 0.5 | 0.25 | 0.5 | 0.5 | 0.9 | 0.92 | 0.92 |
| ALH | 1 | 1 | 1 | 1 | 1 | 0.5 | 0.25 | 0.5 | 0.5 | -1 | 0.5 | 0.5 | 0.25 | 0.5 | 0.5 | 0.25 | 0.5 | 1 | 1 | 1 |
| ALC | 1 | 1 | 1 | 1 | 1 | 0.5 | 0.5 | 0.25 | 0.5 | 0.5 | -1 | 0.5 | 0.5 | 0.25 | 0.5 | 0.5 | 0.25 | 1 | 1 | 1 |
| CLE | 0.75 | 1 | 1 | 0.84 | 0.65 | 0.25 | 0.5 | 0.5 | 0.25 | 0.5 | 0.5 | -1 | 0.5 | 0.5 | 0.25 | 0.5 | 0.5 | 0.9 | 0.92 | 0.92 |
| CLH | 1 | 1 | 1 | 0.84 | 0.9 | 0.5 | 0.25 | 0.5 | 0.5 | 0.25 | 0.5 | 0.5 | -1 | 0.5 | 0.5 | 0.25 | 0.5 | 1 | 1 | 1 |
| CLC | 1 | 1 | 1 | 0.84 | 0.9 | 0.5 | 0.5 | 0.25 | 0.5 | 0.5 | 0.25 | 0.5 | 0.5 | -1 | 0.5 | 0.5 | 0.25 | 1 | 1 | 1 |
| WLE | 0.75 | 1 | 1 | 1 | 0.75 | 0.25 | 0.5 | 0.5 | 0.25 | 0.5 | 0.5 | 0.25 | 0.5 | 0.5 | -1 | 0.5 | 0.5 | 0.9 | 0.92 | 0.92 |
| WLH | 1 | 1 | 1 | 1 | 1 | 0.5 | 0.25 | 0.5 | 0.5 | 0.25 | 0.5 | 0.5 | 0.25 | 0.5 | 0.5 | -1 | 0.5 | 1 | 1 | 1 |
| WLC | 1 | 1 | 1 | 1 | 1 | 0.5 | 0.5 | 0.25 | 0.5 | 0.5 | 0.25 | 0.5 | 0.5 | 0.25 | 0.5 | 0.5 | -1 | 1 | 1 | 1 |
| SPS | 0.6 | 0.8 | 0.86 | 1 | 0.76 | 0.9 | 1 | 1 | 0.9 | 1 | 1 | 0.9 | 1 | 1 | 0.9 | 1 | 1 | -1 | 0.29 | 0.28 |
| SPN | 0.67 | 0.84 | 0.88 | 1 | 0.8 | 0.92 | 1 | 1 | 0.92 | 1 | 1 | 0.92 | 1 | 1 | 0.92 | 1 | 1 | 0.29 | -1 | 0.02 |
| SPW | 0.67 | 0.84 | 0.88 | 1 | 0.8 | 0.92 | 1 | 1 | 0.91 | 1 | 1 | 0.92 | 1 | 1 | 0.92 | 1 | 1 | 0.28 | 0.02 | -1 |

**Fig. 9.** The AIDM matrix

In this part, the result of the AIDM matrix is analyzed: The results of the linkage family algorithms are participated in the final result based on their Independency degrees. According to Figure 9, the differences between the Independency degrees of these algorithms are 0.25 or 0. 5. The differences are based on the problem solving mechanisms of the algorithms and the distance metrics. Also, k-means is considered independent where the linkages don't use the Euclidean distance metric. On the other hand, since the spectral algorithms use k-means to generate the final results after laplacian transformation, the Independency degrees of the spectral algorithms toward k-means is considered special.

As mentioned earlier, the results of the proposed method are compared with well-known base algorithms such as K-means and Spectral, as well as MCLA (Strehl and Ghosh, 2002), EAC (Fred and Jain, 2005), MAX (Alizadeh et al., 2011), APMM (Alizadeh et al., 2014; Alizadeh et al., 2012), D&I (Yousefnezhad et al., 2013), and WOCCE (Alizadeh et al., 2015) which are the state-of-the-art cluster ensemble (selection) methods. All of these algorithms are implemented in the MATLAB R2014b (8.4) by authors in order to generate experimental results. All results are reported by averaging the result of 10 independent runs of the algorithms which are used in the experiment. In this paper, dT is chosen such that each proposed algorithm reaches to a running time of approximately 2 min on a PC with a certain specifications[8]. The experimental results are given in Table 4. The results are in a form of accuracy (percentage) ± standard deviation which is achieved based on the 10 times running of each algorithm. The best results on each data set are bolded.

Table 4 The accuracies (in percentage) along with the standard deviations achieved in the experiments based on the 10 times running of each algorithm.

| UCI Data sets | Basic Clustering Algorithms | | Cluster Ensemble | | Cluster Ensemble Selection | | | | The Proposed Method | |
|---|---|---|---|---|---|---|---|---|---|---|
| | K-means | Spectral | MCLA | EAC | MAX | APMM | D&I | WOCCE | dT | |
| Arcene | 55.24±0.91 | 58.31±0.22 | 64.68±0.711 | 61.79±0.813 | 62.1±0.238 | 66.28±0.216 | 64.9±0.72 | 65.16±0.32 | 0.34 | **68.13±0.82** |
| Balance Scale | 40.32±2.17 | 49.21±0.87 | 51.36±0.71 | 52±0.28 | 52.1±0.91 | 52.65±0.63 | 54.92±0.42 | 54.88±0.61 | 0.06 | **58.64±0.58** |
| Breast Cancer | 93.7±1.94 | 94.88±0.14 | 96.05±0.52 | 95.02±0.39 | 75.72±0.4 | 96.04±0.88 | 96.37±0.29 | 96.92±0.77 | 0.05 | **98±0.14** |
| Bupa | 54.49±0.98 | 56.72±0.18 | 55.36±0.17 | 55.18±0.51 | 56.17±0.28 | 55.07±0.28 | 56.32±0.91 | 57.02±0.46 | 0.07 | **60.83±0.12** |
| CNAE-9 | 64.9±1.53 | 65.32±0.43 | 78.06±0.532 | 74.84±0.193 | 78.63±0.799 | 77.42±0.792 | 77.8±0.47 | 79.2±0.579 | 0.23 | **81.12±0.459** |
| Galaxy | 30.03±0.48 | 31.24±0.67 | 28.48±0.42 | 31.95±0.92 | 32.78±0.69 | 33.72±0.36 | 34.61±0.52 | 35.88±0.81 | 0.15 | **37.18±0.67** |
| Glass | 42.05±1.37 | 45.78±0.87 | 51.4±0.85 | 45.93±0.63 | 44.17±0.47 | 47.19±0.21 | 50.74±0.34 | **51.82±0.92** | 0.06 | 51±0.78 |
| Half Ring | 75.75±1.24 | 80.61±0.15 | 74.5±0.38 | 77.17±0.92 | 78.48±0.59 | 80±0.42 | 86.38±0.92 | 87.2±0.14 | 0.21 | **97.8±0.31** |
| Ionosphere | 65.51±1.342 | 69.71±0.67 | 71.22±0.21 | 67.8±1.118 | 64.48±0.914 | 70.94±0.13 | 69.21±0.74 | 70.52±0.132 | 0.65 | **73.67±0.341** |
| Iris | 65.3±1.46 | 83.45±0.82 | 89.34±0.66 | 96±1.93 | 72.89±0.32 | 74.11±0.25 | 89.43±0.72 | 92±0.59 | 0.13 | **97.3±0.62** |
| Optdigit | 47.23±0.241 | 54.19±0.45 | 77.15±0.452 | 48.12±0.503 | 76.11±0.65 | 77.1±0.841 | 77.59±0.69 | 77.16±0.21 | 0.15 | **78.56±0.692** |
| Pendigits | 40.97±1.69 | 53.94±0.25 | 58.62±0.73 | 43.9±0.43 | 57.02±0.521 | 47.4±0.699 | 59.87±0.81 | 58.68±0.18 | 0.12 | **64.13±0.42** |
| SA Hart | 64.51±0.74 | 69.59±0.08 | 62.54±0.57 | 65.19±1.92 | 63.96±0.72 | 70.91±0.42 | 67.99±0.74 | 68.7±0.46 | 0.5 | **69.05±0.16** |
| Sonar | 51.28±0.32 | 53.24±0.62 | 53.87±0.521 | 52.07±0.651 | 53.98±0.16 | 54.1±0.91 | 53.19±0.52 | 54.39±0.25 | 0.66 | **58.06±0.873** |
| Statlog | 40.89±1.831 | 42.87±0.62 | 55.71±0.342 | 43.96±0.817 | 54.23±0.14 | 54.88±0.528 | 55.46±0.18 | 55.77±0.719 | 0.34 | **57.76±0.591** |
| Wine | 65.73±0.51 | 73.09±0.38 | 70.22±0.33 | 70.56±0.89 | 69.17±0.789 | 64.6±0.231 | 70.19±0.24 | 71.34±0.542 | 0.53 | **74.46±0.141** |
| Yeast | 31.19±0.692 | **32.96±0.71** | 17.56±0.962 | 31.74±0.234 | 32.4±0.124 | 31.06±0.245 | 31.92±0.83 | 32.76±0.268 | 0.1 | 30.12±0.462 |

According to Table 4, although basic clustering algorithms have shown high performance in some data sets, they cannot recognize true patterns in all of them. As mentioned earlier in this paper, in order to solve the clustering problem, each basic algorithm considers a special perspective of a data set which is based on its objective function. The achieved results of basic clustering algorithms which are depicted in Table 4 are good evidences for this claim. Furthermore, the results generated by MCLA and EAC show the effect of the aggregation method on improving accuracy in the final results.

According to Table 4, WOCCE and the proposed algorithm have generated better results in comparison with other basic and ensemble algorithms. Even though the proposed method was outperformed by a number of algorithms in three data sets (Glass, SA Hart and Yaest), the majority of the results demonstrate the superior accuracy of the proposed method in comparison with other algorithms. To accurately clarify the superiority of our proposed method in comparison with its powerful ensemble rivals, the last row of Table 4 (Average) shows the average of accuracy which is achieved in each method. Indeed, as a classic ensemble method, EAC doesn't have any evaluation and selection in its process. This method cannot omit errors which are made in the process of recognizing patterns of the basic clustering results by using the correct information of other basic algorithms' results. The results of EAC which are given in Table 4 show the effects of evaluation and selection in cluster ensemble selection methods.

---
[8] Apple Mac Book Pro, CPU = Intel Core i7(4*2.4 GHz), RAM = 8GB, OS = OS X 10.10

### 4.4. Parameter Analysis

In this section, the effect of diversity threshold on the performance and runtime are analyzed. Hereby, the main goal of our experiment is to show the relation between performance and runtime in the proposed method and to illustrate how the optimized values for the diversity threshold are determined. Thus, this paper employs multiple data sets, two low dimensional data sets (Half Ring, Iris) as well as two high dimensional data sets (Breast Cancer, Wine), for this experiment. Figure 10 illustrates the relationship between the runtime of the proposed method, based on the number of correctly classified samples, and the diversity thresholds. The vertical axis refers to the runtime and the horizontal axis refers to the diversity threshold.

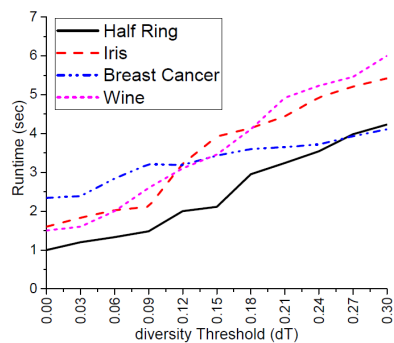

**Fig. 10.** The effect of diversity Threshold (dT) on the runtime of the proposed algorithm

Figure 11 illustrates the relationship between the performance of the proposed method, based on the number of correctly classified samples, and the diversity thresholds. The vertical axis refers to the performance while the horizontal axis refers to the diversity.

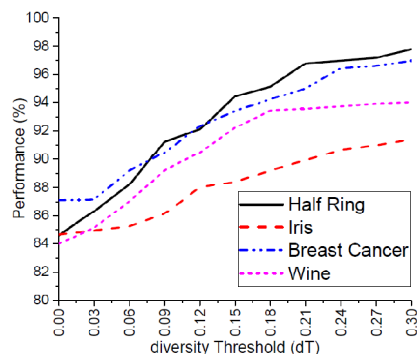

**Fig. 11.** The effect of diversity Threshold (dT) on the performance of the proposed algorithm

As you can see in Figure 10 and Figure 11, although increasing the diversity threshold can improve the performance of the proposed method, it can increase the runtime of the algorithm, too. Therefore, this paper uses a time constant (2 min) to establish a balance between the performance and the runtime.

### 4.5. Noise and Missing-Value Analysis

In this section, a few experiments are conducted in order to analyze the effect of noise and missing values on the performance of the proposed method. This paper employs Arcene and CANE-9 for this experiment, since these two data sets are high-dimensional, large (sample) data sets. Figure 12 illustrates the performance of the proposed method, WOCCE, APMM, MAX, and MCLA on the data sets with missing values. For this cause, some attributes of the mentioned data sets are randomly chosen and their values are set to null. According to Figure 12, WOCCE and the proposed method generate more stable results. As it is clear in this Figure, the proposed method can effectively handle the missing values. The reason is that, it uses the graph based Independency and Uniformity for diversity evaluation.

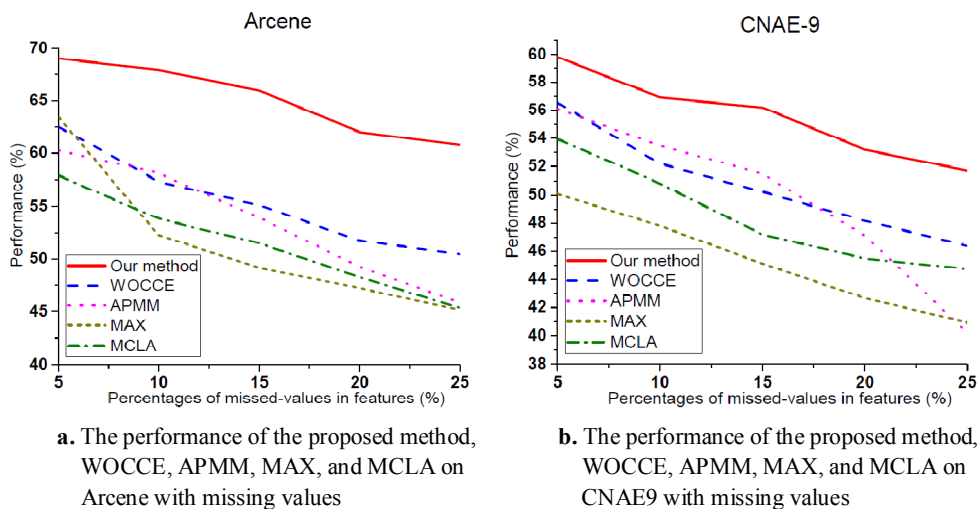

**a.** The performance of the proposed method, WOCCE, APMM, MAX, and MCLA on Arcene with missing values

**b.** The performance of the proposed method, WOCCE, APMM, MAX, and MCLA on CNAE9 with missing values

**Fig. 12.** Missing-Value Analysis

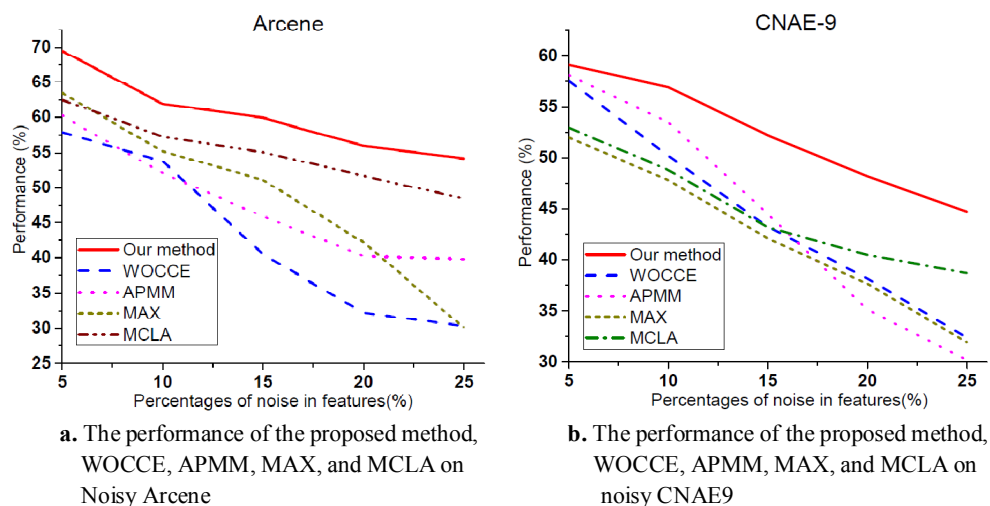

**a.** The performance of the proposed method, WOCCE, APMM, MAX, and MCLA on Noisy Arcene

**b.** The performance of the proposed method, WOCCE, APMM, MAX, and MCLA on noisy CNAE9

**Fig. 13.** Noise Analysis

Figure 13 illustrates the performance of the proposed method, WOCCE, APMM, MAX, and MCLA on the data sets which contain noises. For this cause, some attributes of the mentioned data sets are randomly changed. According to Figure 13, WOCCE and the proposed method generate more stable results. It was claimed earlier that the goal of Independency is to achieve high-performance as well as generating robust and stable results. This experiment can be the best evidence for the mentioned claim.

5. Conclusion

Traditional cluster ensemble methods concentrate on the diversity and quality of the basic results. This paper suggests a new method for employing the graph-based modeling, which is a concept in software testing, for evaluation of basic clustering algorithm's Independency in the cluster ensemble selection. The most important advantage of this employment is the addition of new aspects, such as Independency, which is based on the graph of clustering algorithms, as well as a new framework for selecting high quality basic clustering results. The degree of Independency which is obtained from the proposed method is used as a weight to evaluate diversity in the processes of generating the final result. Also, this paper proposes a procedure to assess the Independency of the base algorithms. This procedure is based on the CAIL, which is a new modeling language for calculating the Independency of clustering algorithms. We also introduce the Uniformity criterion to measure the diversity of the basic results.

To prove the claims of this paper, the results of the proposed method are compared with the results of basic clustering methods, cluster ensemble methods, and cluster ensemble selection methods. The results were achieved by applying the mentioned methods on 17 standard data sets primarily taken from the UCI repository. In our experiment, data sets with different scales (small, average, and large) were used so that the accuracy could be evaluated regardless of the scale of a data set. In addition, in order to be ensured about the accuracy of all results, the experiment has been repeated 10 times. Similar to other pioneering ideas, the proposed framework can be improved later. This paper suggests employing more basic clustering algorithms in order to better satisfying the diversity in the basic results.